\documentclass[runningheads]{llncs}
\usepackage{xcolor} 
\usepackage{tabularx,ragged2e,booktabs, makecell}

\usepackage{graphicx}
\usepackage{float}
\usepackage{wrapfig} 
\begin{document}
\title{CovidAlert - A Wristwatch-based System to Alert Users from Face Touching}
%
%
\author{Mrinmoy Roy \and
Venkata Devesh Reddy Seethi \and
Rami Lake \and
Pratool Bharti}
\authorrunning{Roy et al.}
%
\institute{Northern Illinois University, Dekalb IL 60115, USA \\
\email{\{mroy, devesh, rlake1, pbharti\}@niu.edu}}
\maketitle              
\begin{abstract}
Worldwide $219$ million people have been infected and $4.5$ million have lost their lives in ongoing Covid-$19$ pandemic. Until vaccines became widely available, precautions and safety measures like wearing masks, physical distancing, avoiding face touching were some of the primary means to curb the spread of virus. Face touching is a compulsive human behavior that can not be prevented without constantly making a conscious effort, even then it is inevitable. To address this problem, we have designed a smartwatch-based solution, CovidAlert, that leverages Random Forest algorithm trained on accelerometer and gyroscope data from the smartwatch to detect hand transition to face and sends a quick haptic alert to the users. CovidAlert is highly energy efficient as it employs STA/LTA algorithm as a gatekeeper to curtail the usage of Random Forest model on the watch when user is inactive. The overall accuracy of system is $88.4\%$ with low false negatives and false positives. We also demonstrated the system viability by implementing it on a commercial Fossil Gen $5$ smartwatch.

\end{abstract}
\section{Introduction}
\label{sec:introduction}
    Compulsive human behaviors and habits such as face touching cause self inoculation of germs/viruses that may lead to the contraction of influenza or a viral disease~\cite{macias2009controlling}. Needless to say, they have the potential to cause severe harm to  individual and public health. As witnessed in the ongoing Coronavirus disease $2019$ (Covid-$19$) pandemic, so far $219$ million people have got infected while $4.5$ million people have lost their lives globally~\cite{whodash}. Up until vaccines were widely available, safety precautions such as wearing masks, maintaining physical distance, and avoiding face touching were the primary means to reduce the spread of virus in the ongoing pandemic~\cite{gudi2020preparedness}. Although a person could make a conscious effort to wear the mask regularly, it is not comfortable to use them continuously for longer periods of time~\cite{health2020desk}. Face touching is a repetitive habit often done unconsciously that makes it difficult to overcome such habit~\cite{bbc2020touch}. Therefore, correction of such behavior requires an intervention tool that is readily accessible, accurate, fast enough to detect face touching in real time, and provides convenience of use.
    
    It is important to remember that face touching is often unanticipated that emphasizes the solution to be pervasive and convenient to use for a longer period \cite{utsw2020touch}. With the recent technological advancements and ever increasing popularity across all age groups, smartwatches are a right fit to base the desired solution. Modern smartwatches are equipped with larger memory, storage, and sensors such as accelerometer and gyroscope, that has made them increasingly capable of handling machine learning applications onboard without requiring the need to interface between a smartphone or a cloud computing system to offload computation-intensive tasks. This opens an avenue to integrate and deploy human activity recognition (HAR)-based algorithms on the wrist that can track the user's hand movement in near real time and raise an alarm.
    A smartwatch can raise an alarm in various forms such as with a visual notification, audio cue, or through the sense of touch. Out of the three possible ways, haptic feedback (through sense of touch) such as vibration can immediately draw the attention of user right before their attempt to touch the face \cite{FaceGuard2021}. A gentle and short vibration is also immune to external noise and lighting conditions \cite{exler2017investigating}, and less distracting to other people around. In this work, we explore the challenges and their solutions in designing a smartwatch-based alarm system that can alert the user right before face touching.

    \section{Challenges and Contributions}
    Although the current work appears like a classic application of human activity recognition (HAR) system, it has its own unique challenges. First and foremost, generally a HAR system requires the activity to be detected once it is performed, but in the current work, touching the face must be detected right before its completion as it may be too late by then. Hence, the system must be designed to detect the activities while in transition. Second, collecting and tagging the sensory data when the activity is in transition requires considerably more complex manual work than completed activities. Third, error in activity detection must be low especially the false negatives (predicting not touching the face when it actually happens) as it can be catastrophic. Even though false positive error is undesirable, it has lesser consequences than false negatives. Fourth, to make this system practical it ought to be very efficient to execute on a resource scarce device without depleting much of their energy.
    
    To overcome these challenges, we have used a pervasive wrist-worn device that beeps and vibrates when the transition of hand to face activity is detected. We used a medical-grade pervasive device, Shimmer~\cite{burns2010shimmer}, to collect the sensory data to train and test our proposed system. Additionally, we exhibited a working demo of our system on a consumer-grade Fossil smartwatch. The major contributions of our work are as follows.
\begin{itemize}
    \item Designed a wristwatch-based alarming system that alerts the user from touching their face. The overall accuracy of our system is $88.4$\% for train-test-split and $70.3$\% for leave-one-out approach. 
    \item Employed a highly efficient STA/LTA algorithm to reduce the system energy consumption significantly by only activating the Random Forest model when the user's hand is in active state.
    \item Prepared a transitional dataset~\cite{roydata21} manually tagged with the ground truth activities. We believe our sensory dataset is first of its kind and will be very useful for the HAR research community.
    \item Explored polynomial interaction of statistical features and reported the most important ones.
    \item Implemented the system on off-the-shelf commercial Fossils smartwatch to validate the practicality. 
\end{itemize}


    
\section{Related Works}

    Human activity recognition (HAR) applications are exceedingly popular in tracking personal health with the evolution of modern commercial smartwatches. Currently, Apple and Android watch users can precisely track various activities, including stand, walk, run and other physical exercises throughout the day. In research, smartwatch-based HAR system have been explored for elderly assistance \cite{lutze2017personal}, detection of self-harming activities in psychiatric facilities \cite{watchdog2018bharti}, distracted driving \cite{distracted2018driving}, smoking activities \cite{badhabit2015shoaib}, speed detection \cite{seethi2020cnn} and more. In this section, we have discussed only recent works related to face touching activities in the context of Covid-$19$ pandemic.
    
    In response to the global health emergency of Covid-$19$ outbreak, several researchers have designed wearable devices based on HAR solutions to reduce the spreading of harmful viruses by preventing face touching activities. Among several recent works, D’Aurizio et al.~\cite{haptic2020aurizio} is able to estimate hand proximity to face and notify the user whenever a face touch movement is detected. They have discussed two different approaches - using only accelerometer and the combination of accelerometer and magnetometer. The study showed that using accelerometer and magnetometer together decreases the false positive rate from $38.1\%$ to $3.2\%$ in face touch detection. Although their solution improves the false positive rate to an impressive $3.2\%$, it requires the user to constantly wear a magnetic necklace that might not be readily accessible/ acceptable to the users limiting its practical usage. Additionally, this work doesn't discuss false negative rates which are much more crucial than false positive rates. While high false positive rate might annoy users by raising false alarms, high false negative rates can be catastrophic as system will not raise alert even in case of face touching event.
    
    Another work in this context was published by Kakaraparthi et al.~\cite{FaceSense2021Kakaraparthi} where they have designed an ear-worn system to detect facial touches. The system uses signals from thermal image and electromyography (EMG) sensors to train a deep learning model and achieved $83.4\%$ accuracy in detecting face touching and $90.1\%$ accuracy in face zone detection. Physiological signal from EMG sensor and thermal features are merged to train a Convolutional Neural Network (CNN) for the classification. While their system has achieved high accuracy, it requires the user to wear it on ear constantly which is not very pervasive or comfortable. Also, the system is reliant on another device for hosting the deep learning models that might introduce communication latency. 
    
    Authors of Sudharsan et al.~\cite{COVIDaway2020Sudharsan} leveraged the combination of four sensors: accelerometer, gyroscope, pressure, and rotation vector for continuous monitoring of arm to detect face touching activity. One class classification models of this study, achieves the highest $0.93$ F1 score using only accelerometer data, and Convolutional Neural Network models achieves $0.89$ F1 score using accelerometer and gyroscope data. Since, their system is trained and tested on only four persons data, it requires performance analysis on larger dataset. Additionally, power consumption is not discussed which is very crucial for the viability of the application. 
    
    In \cite{FaceGuard2021}, Michelin et al. proposed a wearable system that alerts user when they attempt to touch their face. Their system streams data from inertial sensors on a wristband and employs a $1$D-Convolutional Neural Network (CNN) that achieved $92\%$ accuracy to detect face touching activities. Additionally, the authors compared user response times from three sensory feedback modalities: visual, auditory, and vibrotactile and observed that vibrotactile feedback had shortest response time of $427.3$ ms. However, the study did not discuss the challenges of real-life implementation and feasibility of implementing computationally intensive CNN algorithms on resource-constrained wrist-worn devices. 
    
    Overall, we observed in recent works that they are short in at least one of the following categories - pervasiveness of device, discussion over false negative rates, energy-efficient algorithms, real-life implementation, and size of dataset. In this work, we have attempted to fill these gaps by leveraging an efficient STA/LTA~\cite{stalta} and Random Forest~\cite{breiman2001randomforest} algorithms on smartwatch to design a system that can accurately detect the face touching activities while managing energy efficiency.

\subsection{Dataset Preparation}
In this section, we describe different parts of data collection procedure including wearable sensors, sensing modalities, sampling rate, data annotation and recorded relevant activities. Our dataset is publicly made available on GitHub~\cite{roydata21}.

\subsubsection{Wearable Device}
\begin{figure}[ht]
    \centering
    \begin{tabular}{lr}
    \includegraphics[width=0.5\linewidth,height=5cm]{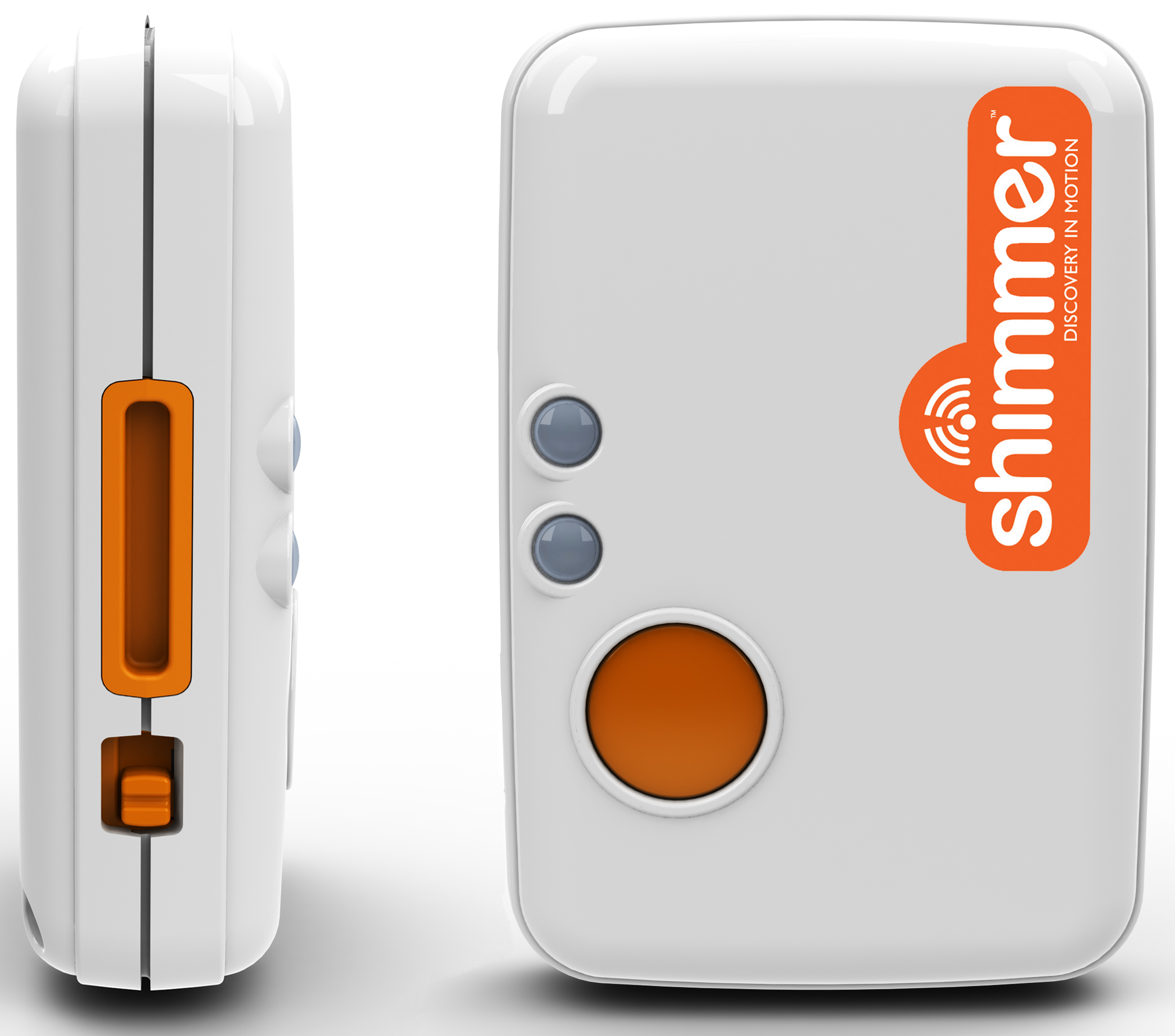} & 
    \includegraphics[width=0.3\linewidth,height=5cm]{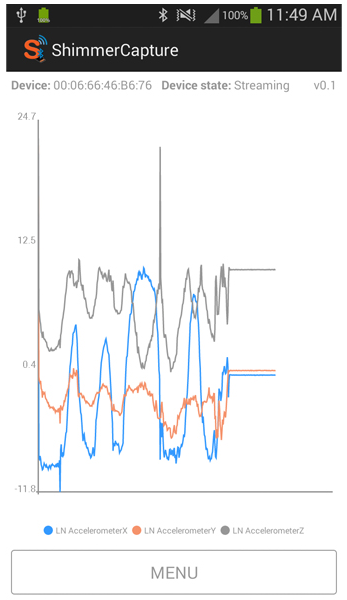}
    \end{tabular}
    \caption{Shimmer device (on left), screenshot of Shimmer Capture Android application (on right).}
    \label{fig:shimmer}
    \end {figure}

We used Shimmer wearable sensing device \cite{burns2010shimmer} for collecting sensory data while tied on the subject's wrist. It has ample processing power along with several multi-modal sensing units. Shimmer is integrated with TI MSP$430$ microcontroller with $24$ MHz CPU and $16$ KB RAM. It contains $11$ channels of $12$ bit A/D with $32$ GB memory resources and $3.7$ V, $450$ mAh re-chargeable lithium polymer battery. It uses class $2$ Bluetooth $2.1$ for live streaming sensory data to a smartphone. The device and the Android application used to capture sensory data is shown in Fig.~\ref{fig:shimmer}.

\subsubsection{Participant Recruitment}
  We obtained permission from the Institutional Review Board at Northern Illinois University (NIU) to facilitate the experiment. We recruited $10$ participants ($2$ female and $8$ male) from NIU. All participants were healthy individuals and gave their written consent before data collection. On average, our participants belonged to an age group of ($34\pm11$ years) and had an average height of ($170\pm12$ cms). Since we conducted the experiment during the ongoing pandemic, we strictly followed the Covid-$19$ protocols such as wearing masks, sanitizing frequently touched spaces, and maintaining social distancing. However, these protocols did not affect the procedures of our study. At the beginning of each data collection session, we secured the Shimmer device on the participant's preferred wrist. Shimmer device has been used extensively in healthcare-related research studies \cite{greene2010quantitative}, \cite{greene2010adaptive}. We leveraged the tri-axial accelerometer (captures acceleration in x-,y-, and z-direction) and tri-axial gyroscope (captures angular velocities in yaw, roll, and pitch directions) from the Shimmer device. The accelerometer sensor in Shimmer cancels the gravitational effects and captures low noise acceleration for the accelerometer. We sampled both sensors with a sampling rate of $102.4$ Hz to obtain high-resolution data. 

\begin{figure}
    \centering
    \includegraphics[width = 0.9\linewidth, keepaspectratio]{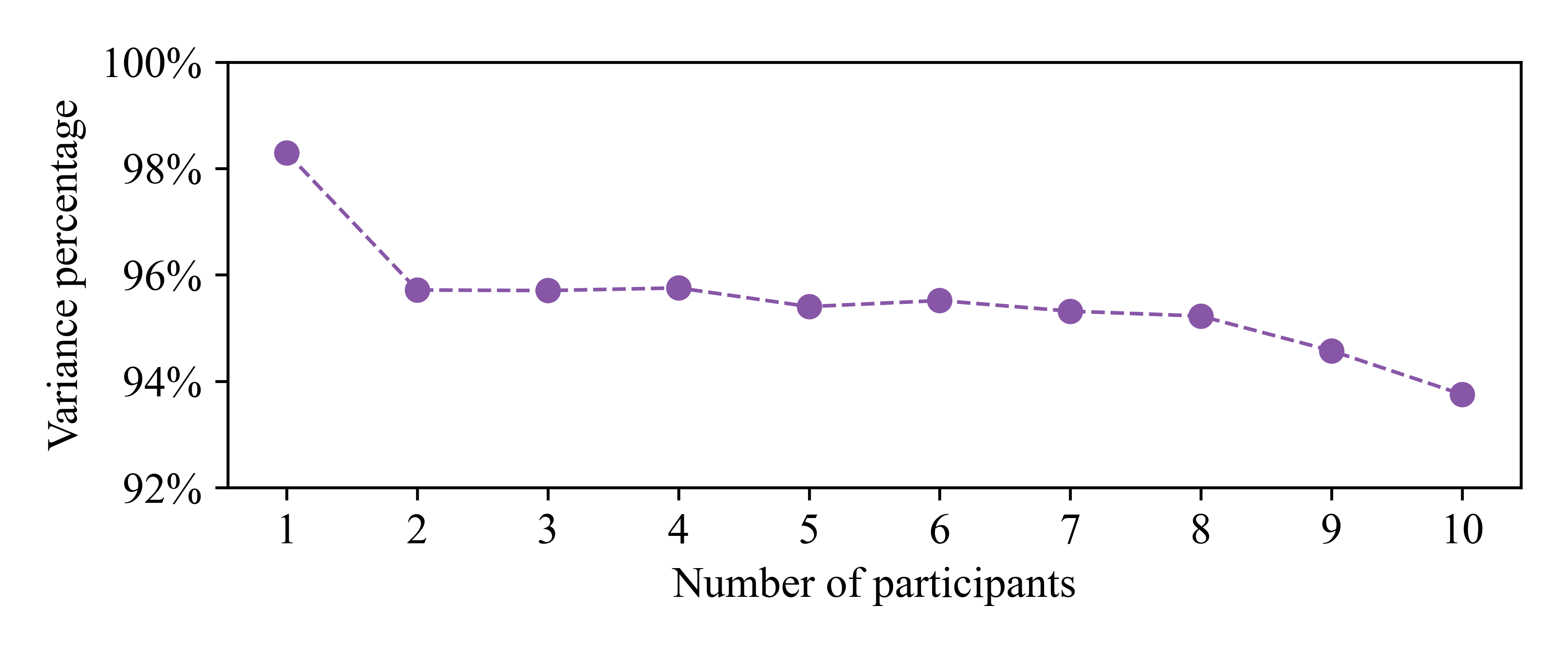}
    \caption{Maximum percentage of variance captured by the first principal component when combining data from different number of participants.}
    \label{fig:variance}
\end{figure}

To estimate if we have collected enough data from participants, we conducted a study based on principal component analysis (PCA)~\cite{wold1987principal} on the $340$ statistical polynomial features generated from accelerometer and gyroscope data (please refer to Section~\ref{sec:extract} for details on feature generation). Briefly, PCA takes features as input and builds a covariance matrix where the eigenvalues represent principal components (PCs). Usually, the higher the variance in dataset, the more the variance is distributed among PCs and the lower the variance percentage captured by the first principal components. We can generate \textit{n} upto number of PCs that are mutually orthogonal from a dataset with $n$ features. We plotted the  Fig.~\ref{fig:variance}  by taking data from $1$ to $10$ participants (on x-axis) and the variance percentage of the first principal component (on y-axis). It was observed that variance percentage decreased from $1$ participant data to $2$ participants data and remained constant up to $8$ participants data. It shows that adding more participants doesn't increase the variance in the dataset significantly. Therefore, we used $10$ participants to capture higher variance as seen in real-life data.      

\vspace{-2em}
\subsubsection{Sensory Data Collection and Annotation}
\label{sec:datacollect}
\begin{figure}[ht]
\centering
\includegraphics[width=0.9\linewidth,keepaspectratio]{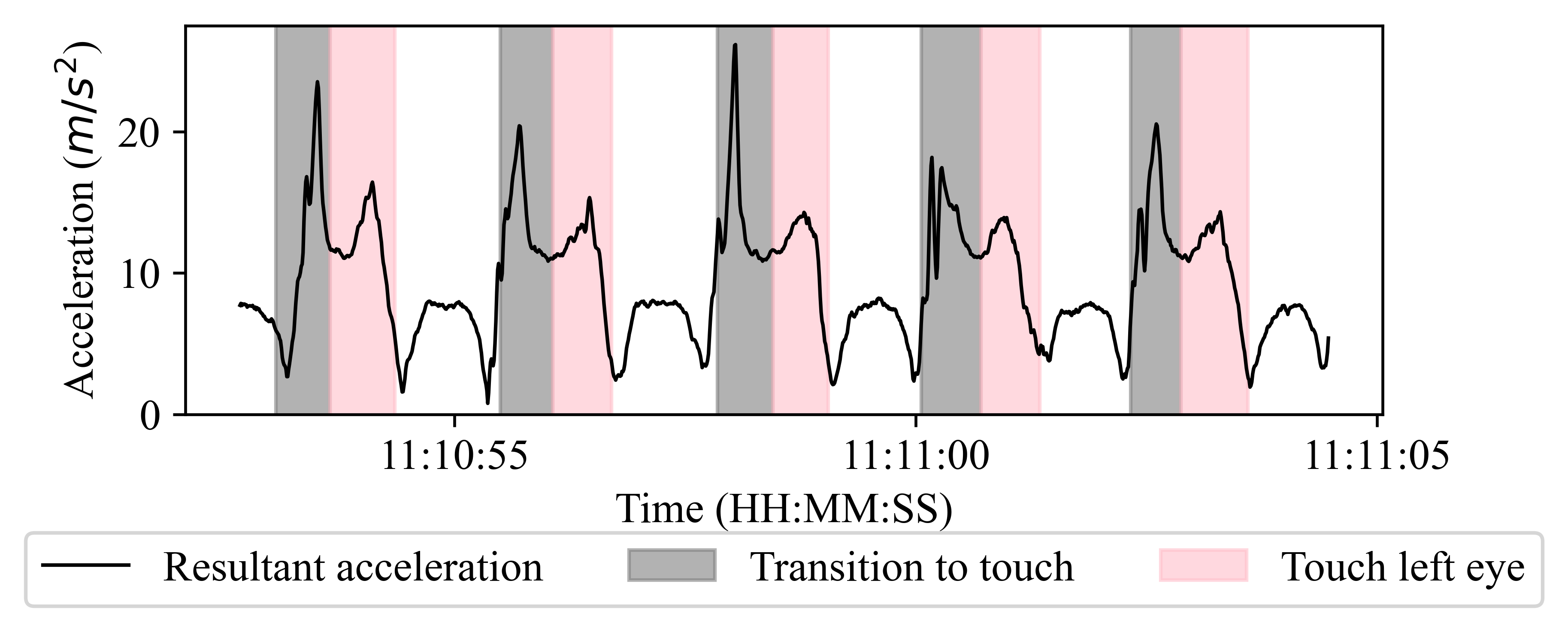}
\caption{Resultant acceleration plotted for touching left eye while standing. The gray and pink hues indicate transition to touch and contact phases, respectively.}
\label{fig:le_sit}
\end {figure}

We collected data from $10$ participants in two categories: face touching and no-face touching. Activities under these two categories are delineated in Table~\ref{tab:activities}. While face touching activities consist of touching different parts of the face such as nose, mouth, left eye, and right eye; the no-face touching category comprised of scratching head, picking up an item on the floor, reaching for a shelf, and maintaining a stance (sitting, standing, and walking). Participants repeated each activity in three stances: sitting on chair, standing, and walking. Hence, each participant participated in $24$ sessions to capture data from $8$ activities across all $3$ stances. In each session, participants repetitively engaged in an activity for at least $30$ seconds or until they repeated same activity $18$ times. Participants were allowed to choose their sequence of activities to complete the $24$ sessions. On average, they took $15$--$20$ minutes to complete $24$ sessions. We encouraged the participants to take breaks during or at the end of each session, and in break time, data collection was paused and resumed soon after. We also asked them to engage in natural activities in any way they wished, such as listening to music, watching a video, or conversing with someone. We ensured that participants acted naturally, resulting in a dataset as close to the real world as possible.
\begin{table}[ht]
    \begin{center}
    \begin{tabular}{||l|l||}
        \hline
        \textbf{No-face touching} & \textbf{Face touching} \\
        \hline\hline
        Scratch head & Touching left eye\\
        Pick item from an overhead shelf & Touching right eye\\
        Pick item from the ground & Touching nose\\
        Stance (sitting, standing, walking) & Touching mouth\\ 
        \hline
    \end{tabular}
    \end{center}
    \label{tab:activities}
    \caption{Face touching and no-face touching activities in our data collection protocol.}
\end{table}
Annotating the sensory data with their ground truth activity was challenging since the main goal is to detect the face touching activity before it gets completed, i.e., in the transition. To do so, we streamed the sensory data from the Shimmer device to an Android smartphone using the Shimmer Capture application~\cite{burns2010shimmer} on the phone using shimmer capture application (as shown in Fig. \ref{fig:shimmer}). We adopted a semi-supervised annotation technique to capture the ground truths. Hence, we developed a Python application to annotate each session on the fly whether the participant was in transition or engaged in the activity. The application takes our annotations and aligns them with the raw data files by matching their timestamps. We present an example for ``touching left eye while standing" activity in Fig. \ref{fig:le_sit}. The figure highlights the regions of interest with gray and pink hues, which denote that the participant's hand is transitioning to touch the left eye and the hand is in contact with the left eye. It is crucial to detect a face touching activity before it happens. Therefore our novel annotation approach makes our dataset superior to the datasets in previous studies that don't have annotations for transition to face. Since our dataset captures the transition to touch and contact phases, researchers can choose to train their models only on transition data. We made our dataset publicly available in GitHub~\cite{roydata21}.

Post data collection, we cleansed the data by discarding $2.5$ seconds of raw data from the start and end of each session. As a next step, we visually inspected all samples by plotting the raw data from the accelerometer and the gyroscope overlayed with transition and contact phases (as shown in Fig. \ref{fig:le_sit}). We promptly deleted samples detected with inconsistencies in annotations. Finally, we extracted only transitional data to train and evaluate our system.

\vspace{-2em}
\vspace{-1em}

\section{Our Proposed Method}
\begin{figure}[htbp]
\centering
\includegraphics[width=1.0\linewidth,keepaspectratio]{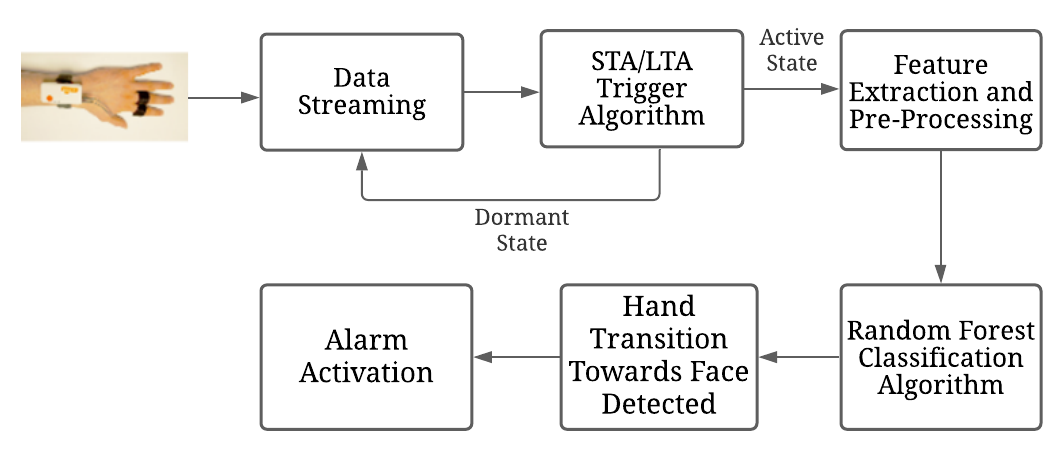}
\caption{System Workflow of CovidAlert.}
\label{fig:flow_face}
\end {figure}

In this section, we introduce the system workflow for CovidAlert and explain the functions of each module in detail. As shown in Fig. \ref{fig:flow_face}, our system streams raw accelerometer and gyroscope data from the Shimmer device and passes it to the STA/LTA triggering algorithm, which acts as a gatekeeper and is in continuous reception mode. The rationale behind STA/LTA algorithm is that when the arm is not moving, it can not touch the face. The triggering algorithm measures the energy of signals using resultant acceleration to determine if the participant's arm is in active or dormant state. If the dormant state is detected, signals are blocked and do not pass forward. However, when an active state is detected, STA/LTA allows the data to pass to Random Forest (RF) module for final classification. If and when the trained RF algorithm anticipates that the user's hand is transitioning to face touch, system alerts the user with the haptic feedback. STA/LTA algorithm is much less computationally extensive than the RF, and effectively it saves significant energy by minimizing the usage of RF algorithm.

\subsection{Data Streaming}
\label{sec:datastreaming}
The role of data streaming module is straight forward. It continuously captures the tri-axial accelerometer and gyroscope sensory data from Shimmer device and passes it to the STA/LTA triggering algorithm. The sampling frequency of both sensors are fixed at $102.4$ Hz.

\subsection{STA/LTA Triggering Algorithm}
\label{sec:stalta}
One of the biggest challenges in implementing any sophisticated ML solutions on smartwatch is its excessive energy consumption \cite{weiss2016smartwatch}. If we simply implement our ML-based solution on the smartwatch, it will most likely deplete the battery in $3$--$4$ hours, making the solution impractical for real-life usage. To make the solution viable, we employed Short-Time-Average/ Long-Time-Average (STA/LTA) algorithm as a gatekeeper in our system which selectively decides when to use the ML algorithm instead of always utilizing it. The motive behind STA/LTA is to save the smartwatch energy by not using ML models when the person's arm is in dormant state. It determines the active/ dormant state by using short time and long time windows. STA/LTA algorithm is used in seismology to detect the intensity of earthquakes \cite{stalta}. In healthcare application, Bharti et al. \cite{watchdog2018bharti} have leveraged the algorithm to detect self-harm activity in psychiatric facilities. 

Taking a real-life scenario where a student attends a lecture, as long as the student is seated with no hand movements, the STA/LTA does not trigger. But, as soon as the student moves their hand to touch the face, the STA/LTA triggers an active state and opens the gate for the ML algorithm for activity recognition. In this manner, STA/LTA module regulates the usage of computationally intensive activity recognition modules and optimizes our CovidAlert to be highly energy-efficient for resource-scarce smartwatches. 

 The algorithm takes resultant acceleration ($A_{xyz}$) as input and calculates the mean ($\mu$) of acceleration, over the Long Term Window ($T_{lta}$) as $L_a = \mu(A_{res}[T_{lta}])$. Similarly, the mean of acceleration, for the Short Term Window, ($T_{sta}$) as $S_a = \mu(A_{res}[T_{sta}])$. When a person's arm remains dormant, both long and short term window has similar energy, hence the ratio typically ranges close to $1$. On the other hand, if the person's arm move suddenly, energy of short term window increases compared to long term window. This means $S_a>>L_a$ or $S_a/L_a>>1$. When the ratio becomes greater than $1$, the triggering algorithm identifies that person is in active state and allows the signals to pass on ML algorithms for final classification. While, length of long term window ranges around $30$--$60$ seconds, typically short term window length is $0.5$--$1$ second.
 

\subsection{Preprocessing and Feature Extraction}
\label{sec:extract}
In this module, we process the raw data to adapt it for machine learning classifier in three stages: 1) segmenting the raw data in sliding windows 2) computing relevant statistical features on segmented windows that have enough discriminative information to classify face/ no-face touching activities 3) generating additional polynomial features from the interaction of features computed in stage $2$. 

\begin{figure}[htbp]
    \centering
    \includegraphics[width=1.0\linewidth,keepaspectratio]{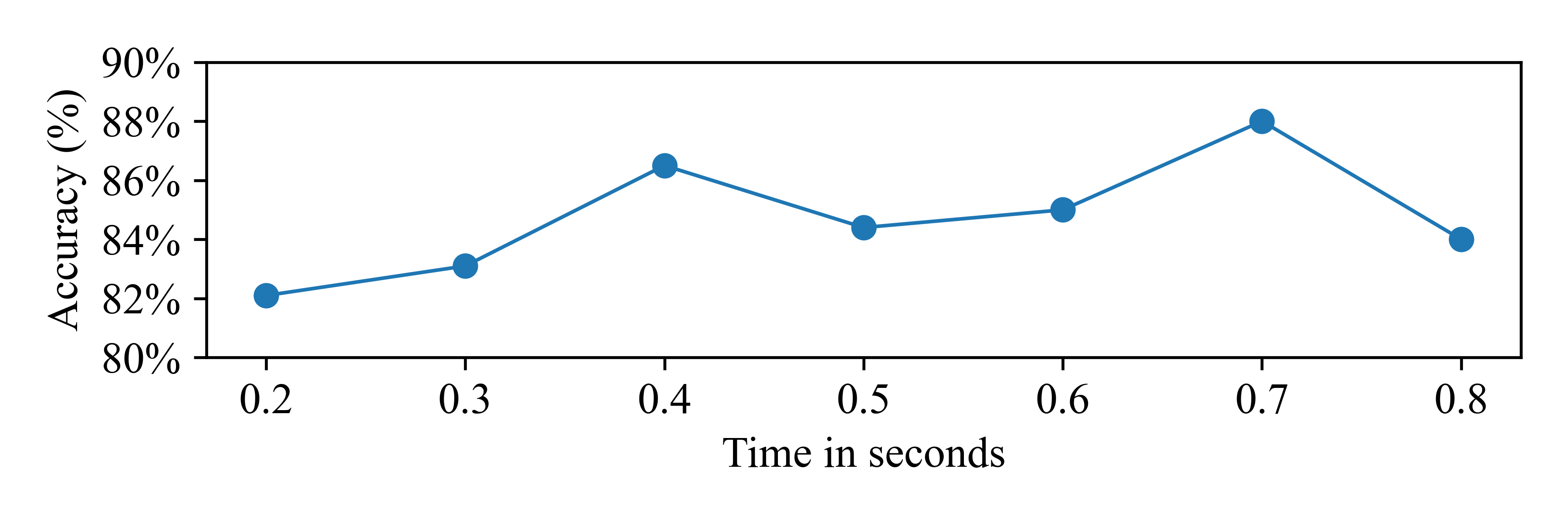}
    \caption{Performance of RF algorithm for window sizes in the range of $0.2$--$0.8$ second.}
    \label{fig:windowSelect}
    \end {figure}

In the first stage, we experimented with different window sizes in the range of $0.2$--$0.8$ seconds. The correct window size is critical in this work because a longer window may fail as users can comfortably touch their faces within one window. Again, if the window is too small, it may not capture the relevant patterns and cause larger false positive errors that may annoy the user. To find the optimal window size, we trained an RF algorithm for different window sizes and reported their accuracies in Fig.~\ref{fig:windowSelect}. We observed the accuracy of RF increased from $82.1\%$ for $0.2$ second window upto $86.5\%$ for $0.4$ second window. The highest accuracy was achieved, $88\%$, for $0.7$ second. However, we still opted for $0.4$ second window as it was large enough to capture slow transitions and small enough to capture swift movements without overlapping with other activities. We then segmented transition parts from each activity in $0.4$ length segments. 
Simply processing raw data from each window requires a large and complex machine learning/ neural network algorithms to uncover the hidden patterns that demands higher computing resources. Smartwatches, however, cannot handle such complex models due to their limited processing and battery resources. Therefore, we chose $9$ simple but relevant statistical features (listed in Table~$2$) that are emblematic to the raw data. Out of the $9$ core features, the first $6$ features are generic for HAR applications features such as minimum, maximum, mean, $25^{th}$ percentile, $75^{th}$ percentile, and standard deviation \cite{casale2011human}. These $6$ features together can effectively communicate the range of values in each axis of the signal. Next, we wanted to use features that represented the qualities of distribution. For this reason, we used skewness and kurtosis that measure the deviation of the data from normal distribution and the tailedness of the data. These two features together give us information about outliers in the given sample and the presence of sudden movements. Finally, we incorporated auto correlation sequence as a feature which finds the correlation with a given vector itself. The correlation score can be used to identify static activities such as standing and sitting and periodic activities such as walking. These statistical features were computed for the three axes of two sensors i.e., accelerometer and gyroscope that generated $9\times 2\times 3= 54$ features. 

\setlength{\extrarowheight}{6pt}
\begin{table}[h!t]
    \begin{center}
    \begin{tabular}{||l|l||}
        \hline
        \textbf{Feature} & \textbf{Description} \\
        \hline\hline
        Minimum value, maximum value, mean  & min($x$), max($x$), $\mu$= mean($x$)\\ \hline
        $25^{th}$ percentile & \makecell[l]{$Q1$ = Feature value dividing\\ first and second quartiles}\\ \hline
        $75^{th}$ percentile & \makecell[l]{$Q3$ = Feature value dividing\\ third and fourth quartiles} \\ \hline
        Standard deviation & $\sqrt{\sum{(x_i-\mu)^2}/N}$\\ \hline
        Skewness & $\sum{(x_i-\mu)^3}/N \times \sigma^3$\\ \hline
	    Kurtosis & $\sum{(x_i-\mu)^4}/N \times \sigma^4$\\ \hline
        Auto correlation sequence & \makecell[l]{Correlation of x with delayed\\ sample of x}\\ \hline
    \end{tabular}
    \end{center}
    \label{tab:statFeatures}
    \caption{Nine statistical features and their descriptions.}
\end{table}

We also generated polynomial features that project the feature space into higher dimensions by learning a polynomial function of degree $2$. This generates interactive features by taking pairs of features at a time that helps discover any non-linear interactions occurring in different features. As a result, we obtained $1540$ polynomial features from the $54$ statistical features.  Finally, we ranked the features according to random forest feature importance and selected a smaller subset of features with comparable accuracy. 


\subsection{Random Forest (RF) Classifier}
\label{sec:rf}

\begin{figure}[htp]
\centering
\includegraphics[width=\linewidth]{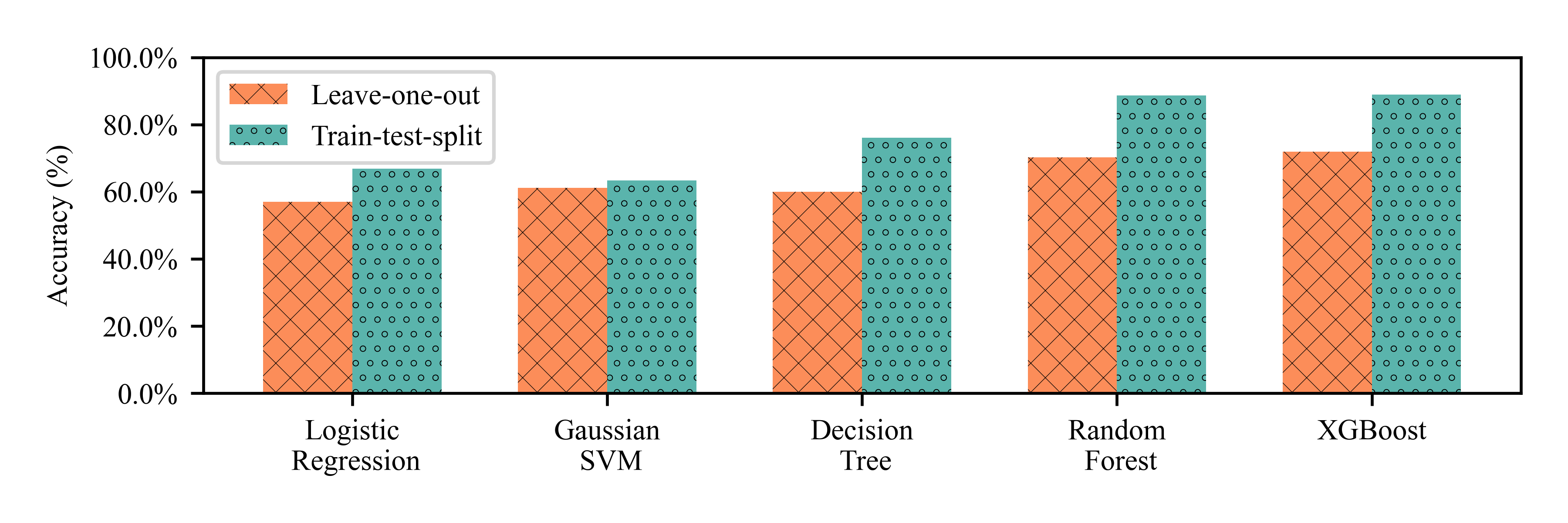}
\caption{Accuracy for different algorithms using leave-one-out and train-test-split evaluation strategies.}
\label{fig:evaluation}
\end {figure}

We first experimented with different machine learning algorithms with all $1540$ polynomial features. We evaluated each algorithm using $80$:$20$ train-test-split and leave-one-out strategies. In train-test-split, $80\%$ data is used in training the model and $20\%$ for evaluation. On the other hand, leave-one-out selects one participant at a time for testing and remaining nine participants data for training and repeats this process for all ten participants in our dataset. The final accuracy score for leave-one-out is the mean of accuracies for all ten participants. We present the accuracies for logistic regression, gaussian support vector machine, decision tree, random forest (RF), and extreme gradient boosting (XGBoost) in Fig. \ref{fig:evaluation}. Train-test-split and leave-one-out metrics are shown in Fig.~\ref{fig:evaluation}. While both RF and XGBoost exhibit superior performance than other classifiers, RF is easier to train and faster compare to XGBoost, hence we picked RF for this study.


RF is one of the simplest but powerful machine learning algorithm. It is an efficient ensemble learning algorithm widely used in numerous recent research studies and real-life applications \cite{oshiro2012many}. RF is an extension of the bagging method that combines several randomized decision trees to create a forest of trees. Each tree in the forest completes the prediction task individually and the predicted class with the most votes are selected as the model prediction. The algorithm is versatile enough to deal with both classification and regression tasks. Since, RF is combination of decision trees, it has a set of hyperparameters to tune for optimizing the performance such as number of trees, maximum number of features used for single tree, minimum number of samples required for a leaf, depth of tree and more. The most important part of training a RF algorithm is to find the optimal hyperparameters which can be done quickly and effectively by using randomized grid search technique \cite{probst2019hyperparameters}.

In this work, we trained an RF algorithm to classify between face/no-face touching activities. First we divided our dataset into training and testing where $80\%$ data is used in training the model and $20\%$ for evaluation purpose. Then, $5$ fold cross validation with randomized grid search is applied for finding the best set of parameters having the best performance score. The optimal parameters were found as bootstrap random sampling = False, maximum depth of tree = $10$, minimum samples required for a leaf = $5$,  minimum samples required to split a node = $20$ and number of trees in a forest = $150$. We also extracted feature importance learned by RF to measure the contribution of each feature while making the final decision. 
\begin{table}[h!t]
    \begin{center}
    \begin{tabular}{||l|c||}
        \hline
        \textbf{Hyperparameter} & \textbf{Value} \\
        \hline\hline
        Maximum depth of tree   &   $10$\\ \hline
        Minimum samples per leaf  &   $5$\\ \hline
        Minimum samples to split a node   &   $20$\\ \hline
        Number of trees in forest    &   $150$\\ \hline
        Bootstrap random sampling   &   False\\ \hline
    \end{tabular}
    \end{center}
    \label{tab:hyperparameters}
    \caption{Selected hyperparameters for RF using randomized grid search.}
\end{table}
\vspace{-3em}
\section{Results}
We trained our RF classifier using the optimal hyperparameters selected from randomized grid search. These hyperparameters are listed in Table~$4$. We employed two evaluation strategies:  train-test-split with $80\%$--$20\%$ train to test split ratio and leave-one-out where repeatedly one participant data is left for the test. In train-test-split, our dataset of $4271$ records ($2080$ no-face touch, $2191$ face touch) is split into $3416$ training records ($1675$ no-face touching, $1741$ face touching) and $855$ testing records ($405$ no-face touching, $450$ face touching). The RF accuracy on the testing dataset using all $1540$ features is $88.7\%$. 

\begin{figure}[htbp]
    \centering
    \includegraphics[width=1.0\linewidth,keepaspectratio]{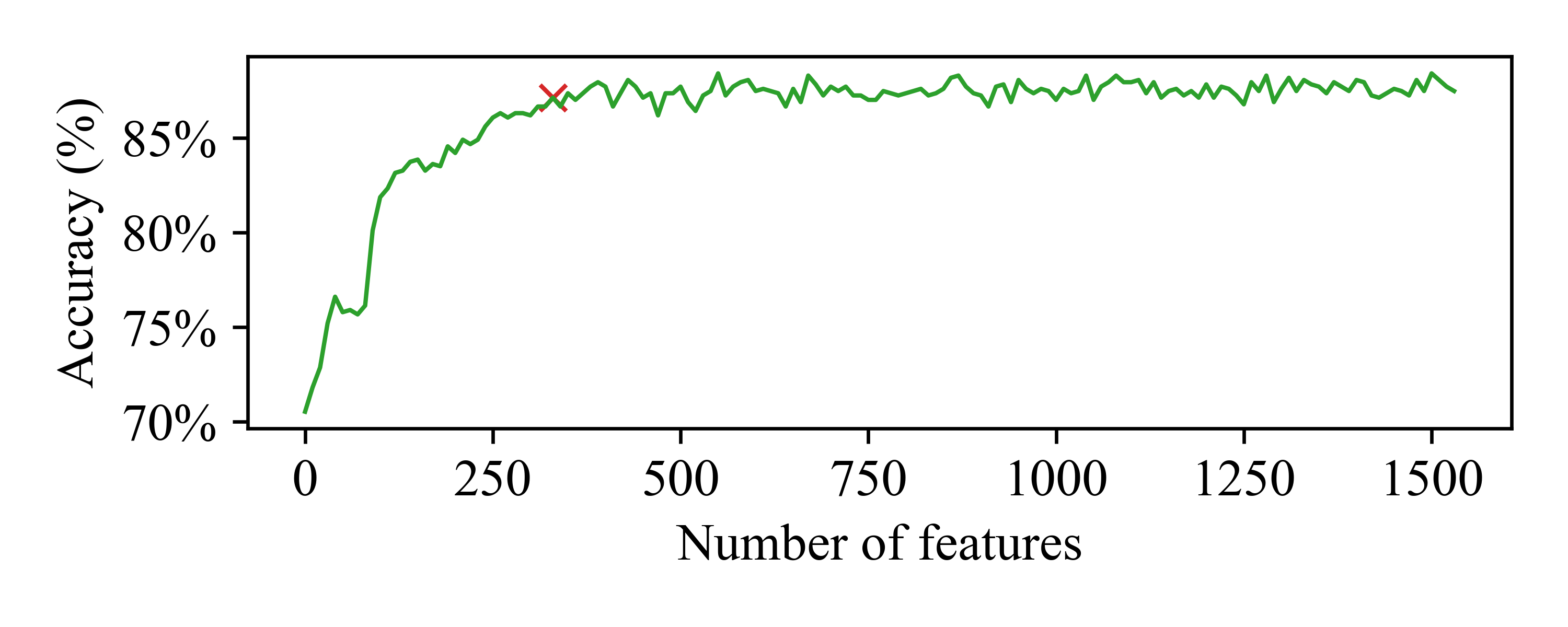}
    \caption{Accuracy of RF taking the features sorted based on their importance scores in the increments of $10$. The red cross was picked using elbow method which indicates that the top $340$ features are sufficient to give good results.}
    \label{fig:featSelect}
    \end {figure}

\begin{figure}[!htbp]
    \centering
    \begin{tabular}{lr}

     \includegraphics[ width=0.5\linewidth,keepaspectratio]{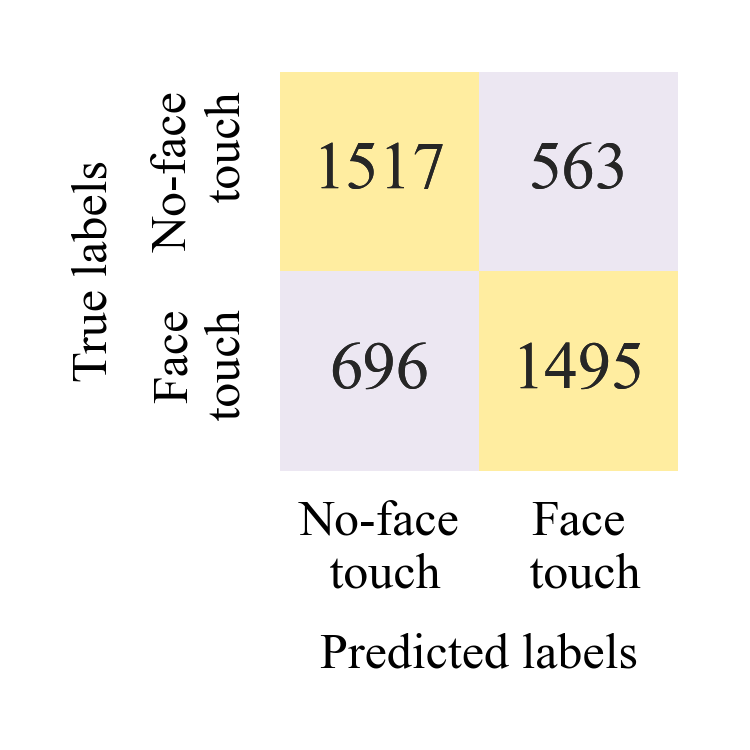} &
     \includegraphics[width=0.5\linewidth,keepaspectratio]{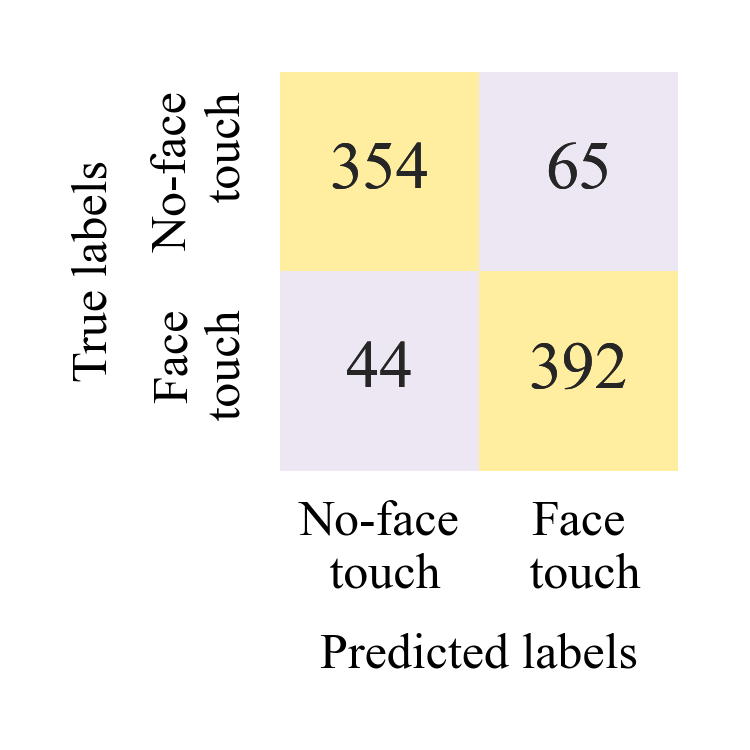} 
    \end{tabular}
    \caption{Confusion matrices for RF trained on $340$ important features using leave-one-out (on left) and train-test-split (on right) strategies.}
    \label{fig:confmat}
    \end {figure}

Although we achieved an impressive result, the model used a large number of features which could be energy intensive to compute especially when deployed on smartwatch. Therefore, to pick minimal number of features that are highly discriminative between the face touch and no-face touch activities, we first ranked all features using RF feature importance~\cite{breiman2001randomforest} and sorted them based on their importance scores. Then, starting from the top $10$ features upto $1540$ features, we iterated in the steps of $10$ features, retrained RF algorithm at each step and cataloged the accuracy score as shown in Fig.~\ref{fig:featSelect}. We found that the RF algorithm by just using the top $340$ achieved $88.4\%$ which was comparable to performance for RF when trained with all $1540$ features. In case of leave-one-out approach, though overall mean accuracy is $70.3\%$, the highest accuracy for a single participant is reported at $87.7\%$. Although accuracy helps us to gauge the algorithm's efficiency, it does not show us the number of false negatives and false positives. As low false negatives are very critical to the application, to measure it, we plotted the confusion matrices in Fig. \ref{fig:confmat} for RF trained on $340$ features for train-test-split and leave-one-out strategies. 

The count of false negatives and false positives were higher in case of leave-one-out than in train-test-split evaluation. However, the number of test samples in leave-one-out evaluation were higher therefore we compare the false positive rate (FPR) and false negative rates (FNR) in both cases. FPR is the ratio of false positives and total negative samples. Similarly, FNR is the ratio of false negatives to total positive samples. In train-test-split, we achieved FPR and FNR of $15.5\%$ and $10\%$ respectively. On the other hand, leave-one-out strategy had FPR and FNR of $27\%$ and $31.7\%$, respectively. As we observe that FPR is higher than FNR for leave-one-out evaluation, this is primarily due to the variance in data from different users.  


    
\section{CovidAlert Demo on Commercial Fossils Watch}
\label{sec:demo}
\begin{figure}[h!tbp]
 \centering
    \includegraphics[width=0.7\linewidth,height=6cm]{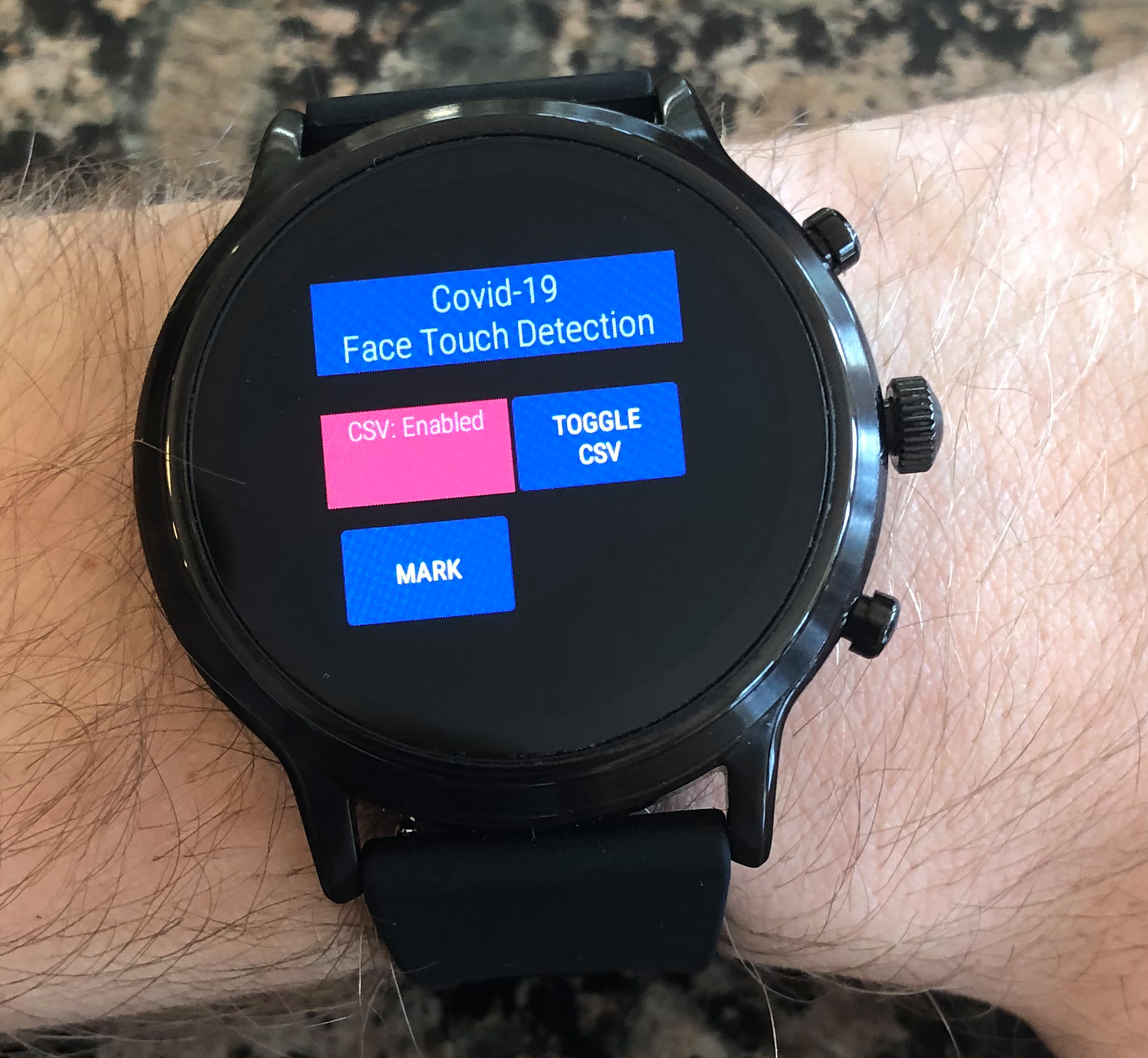}
    \caption{CovidAlert application on Fossil Gen 5 smartwatch.}
    \label{fig:batterytest}
    \end {figure}

 We implemented CovidAlert on a commercial Fossil Gen $5$ smartwatch to validate the practicality of the solution (as shown in Fig. \ref{fig:batterytest}). To do so, we evaluated the duration of time taken to deplete the battery completely in three different scenarios. In all $3$ scenarios, the RF model was working without any STA/ LTA gatekeeping. The first scenario was a control test where the watch was kept on a plain surface without any movement for the complete duration. It took $6$ hours before the battery depleted. In second test, the watch was worn on wrist while doing regular activity that took $4$ hours before battery was depleted. In third test, we disabled the connection to the phone paired to the watch, and enabled airplane mode restricting the watches Wi-Fi, and Bluetooth abilities. These restrictions allowed the application to run for $5$ hours. Simply running STA/LTA algorithm on accelerometer data allows the watch to run for $12$ hours before depleting the battery. Higher usage of STA/LTA will save greater amount of energy by keeping the RF model idle for longer period. 
 
\section{Conclusion and Future Works}
In this paper, we described a smartwatch-based solution to classify transition of hand to face/ no-face touching activities. We employed STA/LTA algorithm to act as a gatekeeper to significantly reduce the usage of computational extensive RF models on the watch by keeping it idle when the user is inactive. Our system is fast in detecting the activity in transition as it requires only $0.4$ second of data to provide its prediction. The overall system accuracy for train-test-split and leave-one-out evaluation strategy is $88.4\%$ and $70.3\%$, respectively, with low false negatives rates. We believe our system is practical and can be used in real life as a safety measure to protect ourselves from self-inoculation of infectious disease like Covid-19. We also implemented CovidAlert on a commercial Fossil Gen 5 smartwatch and discussed its viability. 

In future, we would like to collect feedback from a larger population to improve its practicality. We also want to include more similar kinds of activities to further reduce the false negative and false positive rates. Additionally, we want our system to become smarter by prompting users to wash their hands by automatically detecting activities like a handshake or returning home from outdoors. 

\bibliographystyle{unsrt}
\bibliography{bibfile}
\end{document}